\documentclass[10pt,twocolumn,letterpaper]{article}
\usepackage[accsupp]{axessibility} %
\usepackage{iccv}
\usepackage{times}
\usepackage{epsfig}

\usepackage{bm}
\usepackage{booktabs}
\usepackage{listings}
\usepackage{multirow}
\usepackage{graphicx}
\usepackage[table,xcdraw]{xcolor}
\usepackage{amsthm,amsmath,amssymb}
\usepackage{mathrsfs}
\usepackage{bbding}
\usepackage{lipsum}
\usepackage{arydshln}

\usepackage{algorithmic}
\usepackage[ruled]{algorithm2e}

\usepackage[pagebackref=true,breaklinks=true,letterpaper=true,colorlinks,bookmarks=false]{hyperref}

\iccvfinalcopy 

\def\iccvPaperID{5372} 

\ificcvfinal\fi

\begin{document}

\title{Representation Disparity-aware Distillation for 3D Object Detection}

\author{Yanjing~Li\textsuperscript{1}$^{\dagger}$, 
Sheng~Xu\textsuperscript{1}$^{\dagger}$, 
Mingbao~Lin\textsuperscript{3}, 
Jihao~Yin\textsuperscript{1}, 
Baochang~Zhang\textsuperscript{1,2,4},
Xianbin~Cao\textsuperscript{1}$^{*}$\\
\textsuperscript{1} Beihang University \quad
\textsuperscript{2} Zhongguancun Laboratory\quad
\textsuperscript{3} Tencent\\
\textsuperscript{4} Nanchang Institute of Technology \quad
}
\maketitle
\newcommand\blfootnote[1]{%
\begingroup 
\renewcommand\thefootnote{}\footnotetext{#1}%
\addtocounter{footnote}{0}%
\endgroup 
}
\blfootnote{$^{\dagger}$ Equal contribution.}
\blfootnote{$^*$ Corresponding author: xbcao@buaa.edu.cn}
\blfootnote{$^1$ Code: \url{https://github.com/YanjingLi0202/RDD}}

\maketitle
\ificcvfinal\thispagestyle{empty}\fi

\begin{abstract}
In this paper, we focus on developing knowledge distillation (KD) for compact 3D detectors. We observe that off-the-shelf KD methods manifest their efficacy only when the teacher model and student counterpart share similar intermediate feature representations.  This might explain why they are less effective in building extreme-compact 3D detectors where significant representation disparity arises due primarily to the intrinsic sparsity and irregularity in 3D point clouds. This paper presents a novel representation disparity-aware distillation (RDD) method to address the representation disparity issue and reduce performance gap between compact students and over-parameterized teachers. This is accomplished by building our RDD from an innovative perspective of information bottleneck (IB),  which can effectively minimize the disparity  of   proposal region pairs from student and teacher in features and logits. Extensive experiments are performed to demonstrate the superiority of our RDD over existing KD methods. For example, our RDD increases mAP of CP-Voxel-S to 57.1\% on nuScenes dataset, which even surpasses teacher performance while taking up only 42\% FLOPs. 



\end{abstract}

\section{Introduction}
\label{sec:intro}
3D object detection in point clouds~\cite{qi2017pointnet++,yan2018second,shi2019pointrcnn} is a fundamental perception task with broad applications on autonomous driving, robotics and smart city, \emph{etc}. Beneficial from the large-scale 3D perception datasets~\cite{geiger2012we,caesar2020nuscenes,sun2020scalability} as well as advanced point~\cite{qi2017pointnet++}, pillar~\cite{lang2019pointpillars,wang2020pillar} and voxel based~\cite{graham2017submanifold,liu2019structured} representations of sparse and irregular LiDAR point cloud scenes, 3D detection has achieved remarkable progress~\cite{shi2020pv,yin2021center}. Unfortunately, stronger performance is often accompanied with heavier computation burden, therefore the adoption in real-world applications still remains a challenging problem.

Recent attempts to improve efficiency focus on developing specified architectures for point-based 3D object detectors~\cite{chen2019fast,zhang2022not}, not generalizable to a wide spectrum of pillar/voxel-based methods~\cite{zhou2018voxelnet,lang2019pointpillars,shi2020pv,yin2021center,deng2021voxel,yan2018second}. Here, we aim at a model-agnostic framework for obtaining efficient and accurate 3D object detectors with knowledge distillation (KD). Due to its effectiveness, generality and simplicity, KD has become a popular strategy to develop efficient models in a variety of 2D tasks~\cite{hinton2015distilling,liu2019structured,dai2021general,hou2019learning}, which improves the performance of a lightweight student model by harvesting knowledge from an accurate yet computationally heavy teacher model.

\begin{figure}
\centering
\includegraphics[scale=.3]{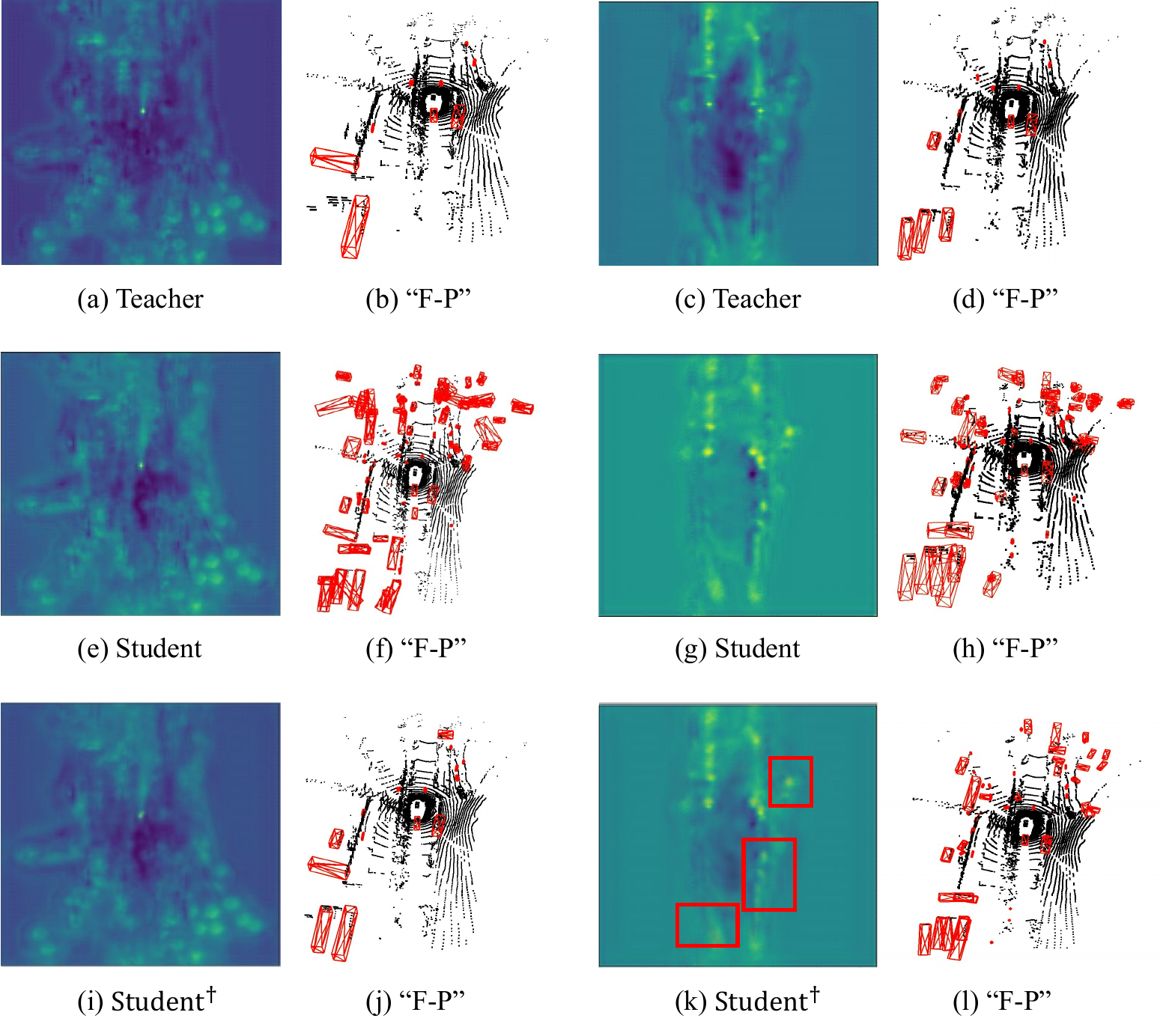}\\
\caption{Visualization of intermediate neck features from teacher CP-Voxel~\cite{yin2021center} and student CP-Voxel-XXS~\cite{yang2022towards}. Student$^{\dag}$ denotes CP-Voxel-XXS distilled by~\cite{yang2022towards}. ``F-P'' denotes false positive predictions from the detector. The second and fourth column images show false positives on the original inputs where the {\color{red} {red}} boxes denote false positives from the detector. Off-the-shelf implementation fails to tackle false positives if significant disparity exists between teacher (c) and student (g) feature maps.}
\label{fig:motivation}
\end{figure}

The recent art~\cite{yang2022towards} employs pivotal position logit KD to enhance the performance of compact 3D detectors. However, as we analyze in this paper, the intrinsic representation disparity, stemming from region distances between compact students and pre-trained teachers, is crucial to 3D student detectors and solely neglected in existing methods~\cite{yang2022towards}.


\begin{figure}
\centering
\includegraphics[scale=.29]{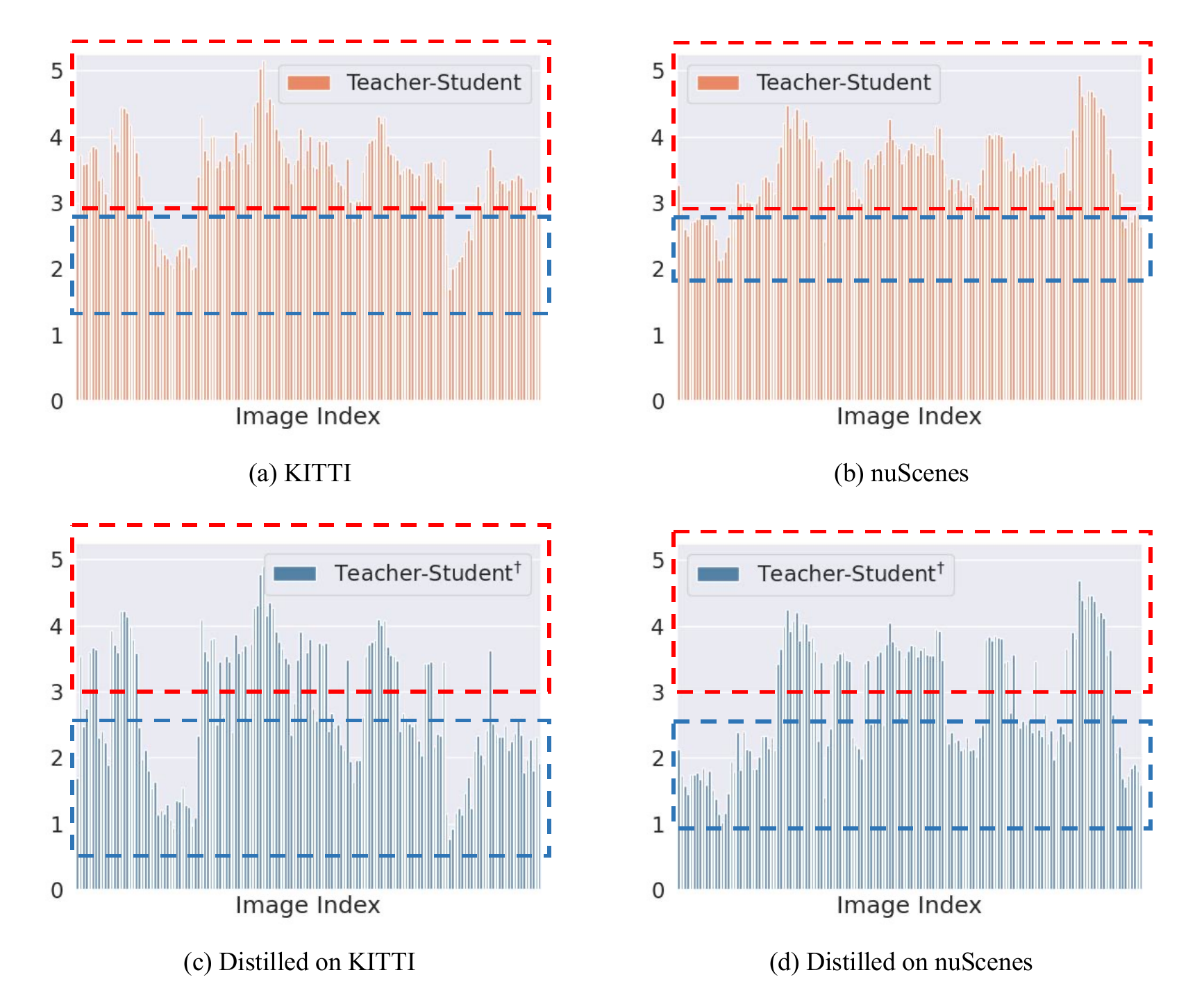}\\
\caption{Histogram of \textit{mean square distance} between feature maps of CP-Voxel $\&$ CP-Voxel-XXS ({\color{orange} {orange}}), and CP-Voxel $\&$ distilled CP-Voxel-XXS-PP ({\color{blue} {blue}}) on KITTI~\cite{geiger2012we} and nuScenes~\cite{caesar2020nuscenes}.}
\label{fig:statistic}
\end{figure}

For an in-depth analysis, in Fig.\,\ref{fig:motivation}, we visualize the feature maps and predictions of pre-trained heavy CP-Voxel~\cite{yin2021center}, a state-of-the-art 3D LiDAR-based detector, and these of compact CP-Voxel-XXS and its distilled version by~\cite{yang2022towards}. The visual results indicate that off-the-shelf KD methods manifest their efficacy only when the teacher model and student counterpart share similar feature maps as in Fig.\,\ref{fig:motivation} $\&$ Fig.\,\ref{fig:motivation}{\color{red}e}. Otherwise, false positives are overmuch if significant representation disparity arises as like Fig.\,\ref{fig:motivation}{\color{red}c} $\&$ Fig.\,\ref{fig:motivation}{\color{red}g}, which greatly deteriorate  the performance of compact 3D detectors. For a comprehensive verification, in Fig.\,\ref{fig:statistic}, we calculate the \textit{mean square distance} between feature maps of teacher CP-Voxel and student CP-Voxel-XXS as a metric to reflect if teacher knowledge can be well transferred to student. It is intuitive that a large distance indicates a higher representation disparity. We perform upon two datasets including KITTI~\cite{geiger2012we} and nuScenes~\cite{caesar2020nuscenes}. Fig.\,\ref{fig:statistic} manifests the statistics where the Fig.\,\ref{fig:statistic} $\&$ Fig.\,\ref{fig:statistic}{\color{red}b} represents distance histogram between teacher and vanilla student while the Fig.\,\ref{fig:statistic}{\color{red}c} $\&$ Fig.\,\ref{fig:statistic}{\color{red}d} represents distance histogram between teacher and distilled student. Each histogram can be separated into 1) the first one in compliance with small distance ({\color{blue} blue area}) and 2) the second one in line with large distance ({\color{orange} orange area}). We observe that the small-distance one is further reduced after distilling, which indicates an efficient distillation. While the large-distance one almost remains unchangeable, which indicates an inefficient distillation. Therefore, it remains an open issue to tackle the representation disparity in existing methods.

\begin{figure*}
\centering
\includegraphics[scale=.34]{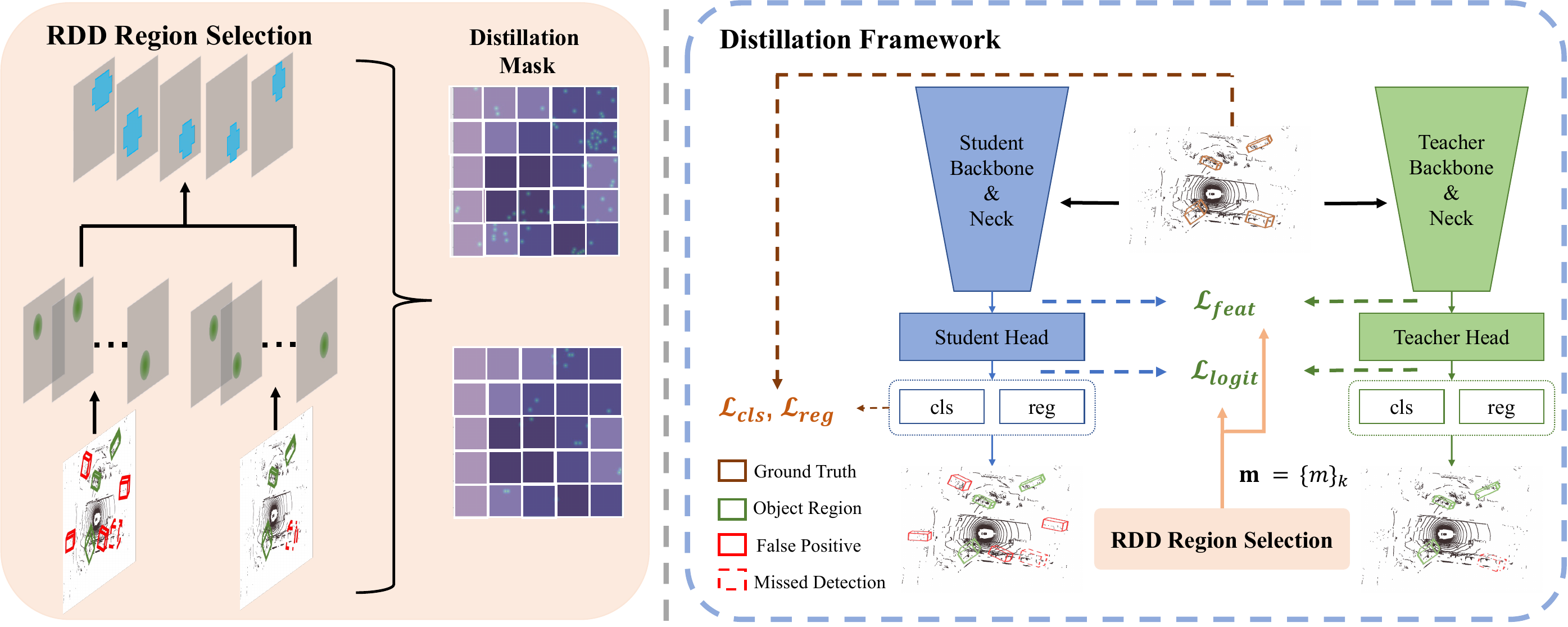}\\
\caption{Overview of the proposed representation disparity-aware distillation (RDD) framework. We first select representative region pairs based on the representation disparity and formulate a balanced mask for each region pair. Then we distill student in both feature-level and logit-level to effectively eliminate the representation disparity.}
\label{fig:framework}
\end{figure*}

Therefore, in this paper we propose a novel 3D detector oriented representation disparity-aware distillation (RDD) method to address the above issue and reduce performance gap between compact students and over-parameterized teachers. Framework of our RDD is illustrated in Fig.\,\ref{fig:framework}, where the distillation objective is actually formulated under the principle of information bottleneck (IB) to maximize the mutual information between intermediate features of teacher and students. To this end, for each region proposal in teacher (student) model, our RDD first pair it by cropping a counterpart region in the same location of student (teacher) model. We measure representation disparity in each pair with mutual information under the IB framework and then learn to weight the region pairs to better bilaterally transfer information between teacher and student. In contrast to off-the-shelf pivotal position logit KD or simply involving ground truths~\cite{wang2019distilling,yang2022towards}, the weighted information is transferred by a feature-level representation disparity-aware distillation as well as logit-level representation disparity-aware distillation loss.

We compare our RDD against state-of-the-art 2D and 3D KD methods~\cite{dai2021general,sun2020distilling,nguyen2022improving,yang2022towards,zhang2022pointdistiller} on datasets of KITTI~\cite{geiger2012we} and large-scale nuScenes~\cite{caesar2020nuscenes}. Extensive results reveal that our method outperforms the others by a considerable margin. For instance, on nuScenes, the CP-Voxel-S~\cite{yang2022towards} distilled by our RDD obtains 57.1\% mAP with only 42\% FLOPs of CP-Voxel~\cite{yin2021center}, achieving a new state-of-the-art. 

\section{Related work}
\label{sec:related}
\noindent\textbf{3D LiDAR-based Object Detection} targets to localize and classify 3D objects from point clouds. Point-based methods~\cite{shi2019pointrcnn,chen2019fast,yang2019std,zhang2022not} take raw point clouds and leverage PointNet++~\cite{qi2017pointnet++} to extract sparse point features and generate point-wise 3D proposals. Pillar-based works~\cite{lang2019pointpillars,wang2020pillar} finish voxelization in bird eye’s view and extract pillar-wise features with PointNet++. Voxel-based methods~\cite{zhou2018voxelnet,yan2018second,shi2020pv} voxelize point clouds and obtain voxel-wise features with 3D sparse convolutional networks, which has become one of the most popular data treatment. Besides, range-based works~\cite{bewley2020range,sun2021rsn} were proposed for long-range and fast detection. Recently, designing efficient 3D detectors has drawn some attentions~\cite{chen2019fast,zhang2022not} with raw point data treatment. In this work, we focus on exploring model-agnostic knowledge distillation methods to boost the performance of lightweight 3D detectors.

\noindent\textbf{Knowledge Distillation} aims to transfer knowledge from a large teacher model to a lightweight student network, which has become a thriving area in efficient deep learning. The simple-yet-common used KD method~\cite{hinton2015distilling} distills knowledge between teacher and student on the output prediction logits. Another line of research proposed to help student’s optimization with hints stored in informative intermediate features from teacher~\cite{romero2014fitnets,huang2017like,komodakis2017paying,heo2019knowledge,jin2019knowledge,chen2021cross}. In 2D detection task, some works attempt distillation techniques~\cite{li2017mimicking,wang2019distilling,dai2021general,qi2021multi,yang2022focal,nguyen2022improving} by emphasizing instance-wise distillation and feature knowledge. Mimic~\cite{li2017mimicking}, FG~\cite{wang2019distilling} and GID~\cite{dai2021general} sample local region features with box proposals or custom indicators for foreground-aware feature imitation. Label KD~\cite{nguyen2022improving} utilizes teacher’s information for label assignment of student. Moreover, recent methods have also been proposed to distill the 3D LiDAR-based detectors. PointDistiller~\cite{zhang2022pointdistiller} captures and makes usage of the semantic information in the local geometric structure of point clouds for compressing the student. \cite{yang2022towards} leverages cues in teacher prediction to determine the important areas for distillation. Nevertheless, existing 3D LiDAR-based KD only leverage the information from well-trained teachers but neglect the distillation-needed areas in the students. In the contrary, in this work, we propose an enhanced 3D detection KD method which takes into consideration representation disparity and effectively transfer the comprehensive information from well-trained teachers to compact students.

\section{Representation Disparity-aware Distillation}
\label{sec:method}
Fig.\,\ref{fig:framework} illustrates the framework of our RDD for 3D object detection. To complete our landscape, in Sec.\,\ref{sec:preliminaries}, we first model our distillation objective under the principle of information bottleneck (IB)~\cite{shwartz2017opening,wang2020bidet,wang2022efficient}. Then, Sec.\,\ref{region_pairs} depicts the generation of region proposal pairs between teacher and student, and representation disparity in each pair. Lastly, we detail our distillation losses including feature-level representation disparity-aware distillation and logit-level representation disparity-aware distillation.




\subsection{Knowledge Distillation Objective}
\label{sec:preliminaries}

Usually, knowledge distillation involves a to-be-trained student detector and a pre-trained teacher detector, and we distinguish them with scripts $\mathcal{S}$ and $\mathcal{T}$, respectively. 
We start with a novel perspective of information bottleneck (IB) principle~\cite{shwartz2017opening} to explore KD in 3D detectors. 
As discussed in~\cite{wang2020bidet}, IB is commensurate with the best compression hypothesis that advocates minimizing data misfitting and model complexity in concert such that the task-irrelevant information can be well diminished in the compressed model for a better performance. 
Considering the facts in 3D object detection that the point clouds are overwhelmingly sparse and the extreme imbalance keeps going between informative instances and redundant background, an efficient information extraction is therefore particularly important.

Given a set of point clouds $X$, KD objective in the IB principle is written as:
\begin{equation}
\begin{aligned}
\mathop{\min}_{\theta_{\bm B}^{\mathcal{S}}, \theta_{\bm D}^{\mathcal{S}}} [I(X; f^{\mathcal{S}}) - \delta I(f^{\mathcal{S}}; y^{GT})] - \beta I(f^{\mathcal{S}}; f^{\mathcal{T}} ), 
\end{aligned}
\label{eq:distill_IB}
\end{equation}
where $f^{\mathcal{T}}$ and $f^{\mathcal{S}}$ are the high-level feature maps from the neck/backbone of the teacher and the student, respectively. $y^{GT}$ denotes ground-truth. 
$\theta_{\bm B}^{\mathcal{S}}$ and $\theta_{\bm D}^{\mathcal{S}}$ are the parameters of backbone and detection part in student respectively. Meanwhile, $\delta, \beta$ are Lagrange multipliers~\cite{shwartz2017opening}. $I()$ returns the mutual information between its two input variables. 
With $I(f^{\mathcal{S}}; y^{GT})$ maximizing the mutual information between the features and ground-truth, the first item $I(X; f^{\mathcal{S}})$ minimizes the mutual information between the point cloud data and the high-level feature maps of student to control the noise introduction. This part can be treated as the original detection loss of the detector~\cite{yin2021center,lang2019pointpillars}. 
With teacher model’s features as guidance, the second item $\beta I(f^{\mathcal{S}}; f^{\mathcal{T}})$ maximizes the mutual information to preserve more teacher information in student.
The collective cooperation between the two items guides student to focus more on beneficial information and less on noise information~\cite{wang2022efficient,wang2020bidet,shwartz2017opening}.

\subsection{Region Pairs with Representation Disparity}\label{region_pairs}

\textbf{Region Pairs}.
We denote $\{ R_{i}^{\mathcal{T}}|(p_{reg,i}^{\mathcal{T}}, p_{cls,i}^{\mathcal{T}})\}_{i=1}^M$ and $\{ R_{i}^{\mathcal{S}}|(p_{reg,i}^{\mathcal{S}}, p_{cls,i}^{\mathcal{S}})\}_{i=M+1}^{M+N}$ as the outputs of teacher and student where the $i$-th region proposal $R^\mathcal{T}_i/R^\mathcal{S}_i \in \mathbb{R}^{C \times H \times W}$ contains two information including a regression coordinate $p_{reg,i}$ to model the proposal position and a classification probability $p_{cls,i}$ to tell the proposal category.
Note that in the center-based 3D detectors, {\em e.g.}, CenterPoints~\cite{yin2021center}, each proposal corresponds to a Gaussian area in the heatmaps, while in the anchor-based 3D detectors, {\em e.g.}, PointPillars~\cite{lang2019pointpillars}, each proposal corresponds to a region in the intermediate features. 

\begin{figure}
\centering
\includegraphics[scale=.36]{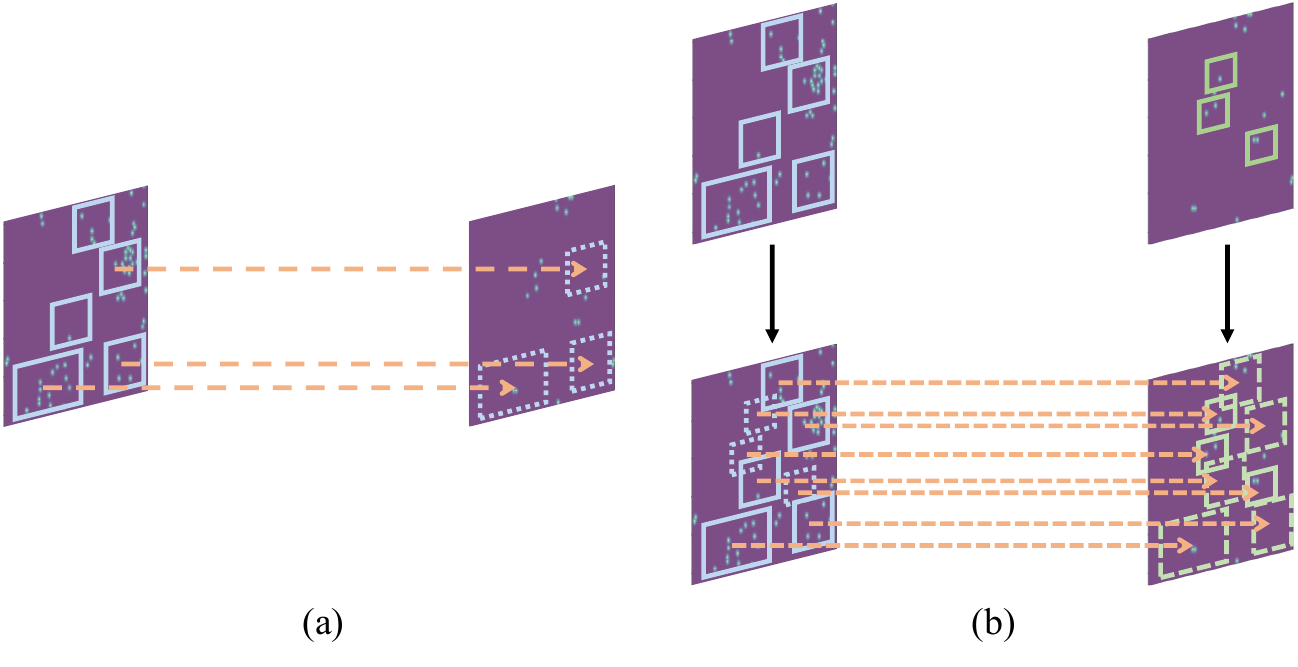}\\
\caption{Illustration for the generation of the region pairs. Each single region in one model generates a counterpart feature map patch in the same location of the other model.}
\label{fig:pair}
\end{figure}

As shown in the left of Fig.\,\ref{fig:pair}, the recent study~\cite{yang2022towards} unilaterally passes on teacher proposals of higher classification probability to the corresponding regions of student. On the one hand, it ignores the efficacy of student information; on the other hand, the representation disparity is neglected as discussed in Sec.\,\ref{sec:intro}. 
In this paper, we propose to bilaterally transfer information between teacher and student. Specifically, as depicted in the right of Fig.\,\ref{fig:pair}, for each single region in teacher (student) model, we crops a counterpart feature map patch in the same location of student (teacher) model to form a total of $M+N$ region pairs 
$\{(R^{\mathcal{T}}_i, R^{\mathcal{S}}_i)\}_{i=1}^{M+N}$. Here, $\{R^{\mathcal{S}}_i\}_{i=1}^{M}$ accords with cropped region patches in student and $\{R^{\mathcal{T}}_i\}_{i=M+1}^{M+N}$ is in tune with these patches in teacher. 
For ease of representation, the superscripts $\mathcal{T}$ and $\mathcal{S}$ will be dropped from time to time in the following contents.

Then, our distillation considers representation disparity issue by weighting the patch-level distance under the IB principle.
Before diving into details, we first channel-wise normalize proposal $R_i$ considering the large-scale magnitude gap between the pre-trained teacher and the to-be-trained student, as:
\begin{equation}
\begin{aligned}
\hat{R}_{i;c,:,:} 
= \frac{\operatorname{exp}(\frac{R_{i;c,:,:}}{\tau})}{\sum_{c' \in \{1,2,...,C\}}\operatorname{exp}(\frac{R_{i;c',:,:}}{\tau})},
\end{aligned}
\label{eq:transform}
\end{equation}
where $R_{i;c,:,:}$ denotes the $c$-th channel of $R_{i}$ and  $\tau=4$ in this paper denotes a hyper-parameter controlling the statistical attributions of the channel-wise alignment operation

\textbf{Representation Disparity}.
Under the framework of IB principle, we define and evaluate representation disparity as mutual information between student patch $\hat{R}^{\mathcal{S}}_i$ and teacher patch $\hat{R}^{\mathcal{T}}_i$. This is formulated as:
\begin{equation}
\begin{split}
I(\hat{R}^{\mathcal{S}}_i; \hat{R}^{\mathcal{T}}_i)
= H(\hat{R}^{\mathcal{S}}_i) - H(\hat{R}^{\mathcal{S}}_i|\hat{R}^{\mathcal{T}}_i),
\end{split}
\label{eq:mutual} 
\end{equation}
where $H()$ returns the information entropy. A smaller $I(\hat{R}^{\mathcal{S}}_i; \hat{R}^{\mathcal{T}}_i)$ indicates higher disparity between $\hat{R}^{\mathcal{S}}_i$ and $\hat{R}^{\mathcal{T}}_i$.

Then, a naive way to measure mutual information between teacher and student $I(f^{\mathcal{S}}; f^{\mathcal{T}})$ in the KD objective of Eq.\,(\ref{eq:distill_IB}) is to sum up mutual information of all patch pairs as:
\begin{equation}
    \begin{aligned}
        I(\hat{R}^{\mathcal{S}}_i; \hat{R}^{\mathcal{T}}_i) = \sum_{i=1}^{M+N}m_iI(\hat{R}^{\mathcal{S}}_i;\hat{R}^{\mathcal{T}}_i). 
    \end{aligned}
\end{equation}

Nevertheless, it does not take into account the representation disparity among different region pairs. Given this, prior to a formal distillation, we propose to optimize a weighting vector ${\bm m}=[m_1,m_2,...,m_{M+N}] \in \mathbb{R}^{M+N}$ to identify disparate region pairs, leading to our learning objective as:
\begin{equation}
\begin{aligned}
\mathop{\min}_{{\bf m}} \underbrace{\sum_{i=1}^{M+N}m_iI(\hat{R}^{\mathcal{S}}_i;\hat{R}^{\mathcal{T}}_i)}_{I(f^{\mathcal{S}}; f^{\mathcal{T}} | {\bm m})} + \lambda \|{\bf m}\|_1, 
\end{aligned}
\label{eq:select_obj}
\end{equation}  
where the minimization leads $m_i \in {\bm m}$ to be large to penalize high disparity stemming from pair $\hat{R}^{\mathcal{S}}_i$ and $\hat{R}^{\mathcal{T}}_i$. The term $\mathop{\min}_{{\bf m}} \|{\bf m}\|_1$ is involved to prevent the model to equally distill all alternative region pairs. Also, to indicate the disparity degree. we clip $m_i$ to $1$ or $0$ if its value is beyond $[0,1]$. $\lambda > 0$ is a hyper-parameter to determine the sparsity of ${\bm m}$.

Moreover, the introduction of ${\bm m}$ leads our format of $I(f^{\mathcal{S}}; f^{\mathcal{T}})$ to $I(f^{\mathcal{S}}; f^{\mathcal{T}} | {\bm m})$ as manifested in Eq.\,(\ref{eq:select_obj}). Together with our representation disparity, the KD objective under IB framework finally changes from Eq.\,(\ref{eq:distill_IB}) to:
\begin{equation}
\begin{aligned}
\mathop{\min}_{\theta_{\bm B}^{\mathcal{S}}, \theta_{\bm D}^{\mathcal{S}}} &[I(X;f^{\mathcal{S}}) - \delta I(f^{\mathcal{S}};y^{GT})] - \beta I(f^{\mathcal{S}}; f^{\mathcal{T}} | {\bm m}^*), 
\\&
\operatorname{s.t.} \; {\bf m}^* = \mathop{\arg \min}_{\bf m} \; I(f^{\mathcal{S}}; f^{\mathcal{T}} | {\bm m}) + \lambda \|{\bm m}\|_1. 
\end{aligned}
\label{eq:final_objective}
\end{equation} 

Our objective involve two sub-problems. In each training iteration, we first perform inner-level optimization to derive a current optimal ${\bm m}^*$; and then solves the upper-level optimization to conduct distillation based on explicit distillation losses in Sec.\,\ref{sec:transfer}. Notice the inner-level optimization causes negligible costs compared to the upper-level one since the size of region pairs is not large. On the contrary, it derives distillation to focus more on disparate region pairs for a better performance.

\subsection{Knowledge Transferring}
\label{sec:transfer}
We present the upper-level optimization in Eq.\,(\ref{eq:final_objective}). Recall in Sec.\,\ref{sec:preliminaries} we analyze that the first item $[I(X;f^{\mathcal{S}}) - \delta I(f^{\mathcal{S}};y^{GT})]$ is in compliance with the original detection loss such as proposal classification and coordinate regression. Our central is to hand over the specific format of the second term $I(f^{\mathcal{S}}; f^{\mathcal{T}} | {\bm m}^*)$, which according to Eq.\,(\ref{eq:mutual}) and Eq.\,(\ref{eq:select_obj}) can be explicitly derived as:
\begin{equation}
I(f^{\mathcal{S}}; f^{\mathcal{T}} | {\bm m}^*) = \sum_{i=1}^{M+N}m^*_i (H(\hat{R}^{\mathcal{S}}_i) - H(\hat{R}^{\mathcal{S}}_i|\hat{R}^{\mathcal{T}}_i)).
\end{equation}

Considering the intrinsic entropy of $\hat{R}^{\mathcal{S}}_i$ remains unchanged within each iteration, therefore $H(\hat{R}^{\mathcal{S}}_i)$ is regarded as a constant. Maximizing $I(f^{\mathcal{S}}; f^{\mathcal{T}} | {\bm m}^*)$ turns to minimizing $H(\hat{R}^{\mathcal{S}}_i|\hat{R}^{\mathcal{T}}_i)$. Unfortunately, it is hard to directly minimize $H(\hat{R}^{\mathcal{S}}_i|\hat{R}^{\mathcal{T}}_i)$. Instead, we choose to minimize norm distance between $\hat{R}^{\mathcal{S}}_i$ and $\hat{R}^{\mathcal{T}}_i$ as a substitute since both of them reach optimal when $ \hat{R}^{\mathcal{S}}_i = \hat{R}^{\mathcal{T}}_i$.

In view of that feature pyramid network (FPN)~\cite{lin2017feature} has been adopted in most 3D detectors for robustness of multi-scale detection~\cite{yin2021center,lang2019pointpillars,yan2018second}, it is natural to choose the neck feature maps after FPN for distillation. After forming the region pairs, the feature-level representation disparity-aware distillation loss is computed as:
\begin{equation}
\begin{aligned}
\mathcal{L}_{feat} &= \frac{1}{M+N} \sum_{i=1}^{M+N} m_i^*  \|\varphi(\psi(\hat{R}_i^{\mathcal{S}})) - \hat{R}_i^{\mathcal{T}} \|_2, 
\end{aligned}
\label{eq:feature_rdd}
\end{equation}
where $\varphi$ indicates the RoI Align~\cite{he2017mask}. $\psi$ is $1 \times 1$ convolution followed by batch normalization~\cite{ioffe2015batch} and ReLU~\cite{nair2010rectified} to align channel-wise discrepancy between teacher region $\hat{R}_i^{\mathcal{T}}$ and and student region $\hat{R}_i^{\mathcal{S}}$~\cite{dai2021general,wang2019distilling}.

In addition to FPN outputs, another neglected special kind of feature maps come to the logits in the classification and regression branches. Previous 2D and 3D methods~\cite{yang2022towards,wang2019distilling,hinton2015distilling} conduct distillation on the whole or pivotal part of the detection head outputs or on the conventional feature maps, degrading the student performance~\cite{dai2021general}. The probable cause can be attributed to the extreme imbalance between instances and backgrounds in 3D LiDAR-based detection task. Therefore, based on the weighting vector ${\bm m}$, we also take into consideration the regression coordinate and classification probability of each region proposal $\hat{R}_i$, and form our logit-level representation disparity-aware distillation loss as:
\begin{equation}
\small
\begin{aligned}
\mathcal{L}_{logit} = \frac{1}{M+N} \sum_{i=1}^{M+N}& m_i^* (\|p_{cls,i}^{\mathcal{S}} - p_{cls,i}^{\mathcal{T}}\|_1 
\\&
+ \| (p_{reg,i}^{\mathcal{S}} - p_{reg,i}^{\mathcal{T}})\|_1),
\end{aligned}
\label{eq:logit_rdd}
\end{equation}
in which $p_{cls, i}$ and $p_{reg,i}$ denote the classification probability and proposal coordinates for the $i$-th region proposal $R_i$ as introduced in Sec.\,\ref{region_pairs}. It should be noted that, for the center-based 3D detector~\cite{yin2021center}, the regression loss item do not exist. 

Finally, the exact definition of our KD objective in Eq.\,(\ref{eq:distill_IB}) is given as:
\begin{equation}
\mathcal{L} = \underbrace{\mathcal{L}_{cls} + \gamma\mathcal{L}_{reg}}_{I(X;f^{\mathcal{S}}) - \delta I(f^{\mathcal{S}};y^{GT})} + \underbrace{\alpha_1\mathcal{L}_{feat} + \alpha_2\mathcal{L}_{logit}}_{-\beta I(f^{\mathcal{S}}; f^{\mathcal{T}})}, 
\label{eq:train_objective}
\end{equation}
where $\mathcal{L}_{cls}$ and $\mathcal{L}_{reg}$ are the original detection losses supervised by the ground-truth labels. The $\gamma, \alpha_1$ and $\alpha_2$ are trade-off parameters between different objectives.


\section{Experiments}
\label{sec:experiments}
Comprehensive experiments are conducted to evaluate our proposed method on two datasets for object detection: nuScenes~\cite{caesar2020nuscenes} and KITTI~\cite{geiger2012we}. First, we introduce the datasets, metrics, implementation details and compact model architecture in Sec.\,\ref{sec:settings}. Then we select the hyper-parameters, validate the effectiveness of the components, and analyze the information of our method through ablation studies in Sec.\,\ref{sec:ablation}. Finally, in Sec.\,\ref{sec:nuScenes} and Sec.\,\ref{sec:kitti}, we compare our method with other image-based 2D distillation methods implemented on 3D detectors, and other 3D distillation methods to demonstrate the superiority of RDD.  

\subsection{Experimental Settings}
\label{sec:settings}
\textbf{Datasets and Evaluation Metrics}. For nuScenes dataset~\cite{caesar2020nuscenes}, metrics are mean average precision (mAP) and the nuScenes detection score (NDS). These metrics are computed in the physical unit. For KITTI dataset~\cite{geiger2012we}, we report the average precision calculated by 40 sampling recall positions for 3D object detection on the validation split. Following the typical protocol, the IoU threshold is set as 0.7 for class Car and 0.5 for class Pedestrians and Cyclists. 

\textbf{Training $\&$ Validation}. For experiments conducted on nuScenes dataset, we follow the same setups as the original CenterPoint \cite{yin2021center}. We use AdamW~\cite{loshchilov2017decoupled} to train the model. The weight decay for AdamW is $1e-2$. Following a cyclic schedule~\cite{smith2017cyclical}, the learning rate is initially $1e-4$ and gradually increased to $1e-3$, and finally decreased to $1e-8$. We train for 20 epochs on 8 NVIDIA Tesla V100 GPUs. During inference, we take the top 100 highest-scored objects as the final predictions. We do not use any post-processing such as non maximum suppression (NMS). We use the toolkit provided with the nuScenes dataset for evaluation.

For experiments conducted on KITTI dataset, we use AdamW~\cite{loshchilov2017decoupled} to train the model. The weight decay for AdamW is $1e - 2$. Following a cyclic schedule, the learning rate is initially $1e - 4$ and gradually increased to $1e - 3$, which is finally decreased to $1e - 8$. We train for 80 epochs on 8 NVIDIA Tesla V100 GPUs. We apply axis aligned non maximum suppression (NMS) with an overlap threshold of 0.5 IoU, following~\cite{lang2019pointpillars} for inference. We evaluate the model performance using the toolkit provided with the KITTI dataset.

\textbf{Compact Model Architecture}.  We apply CenterPoint-Voxel and CenterPoint-Pillar~\cite{yin2021center} on nuScenes, and SECOND~\cite{yan2018second} and PointPillars~\cite{lang2019pointpillars} on KITTI. 

\begin{figure}[t]
\centering
\includegraphics[scale=.36]{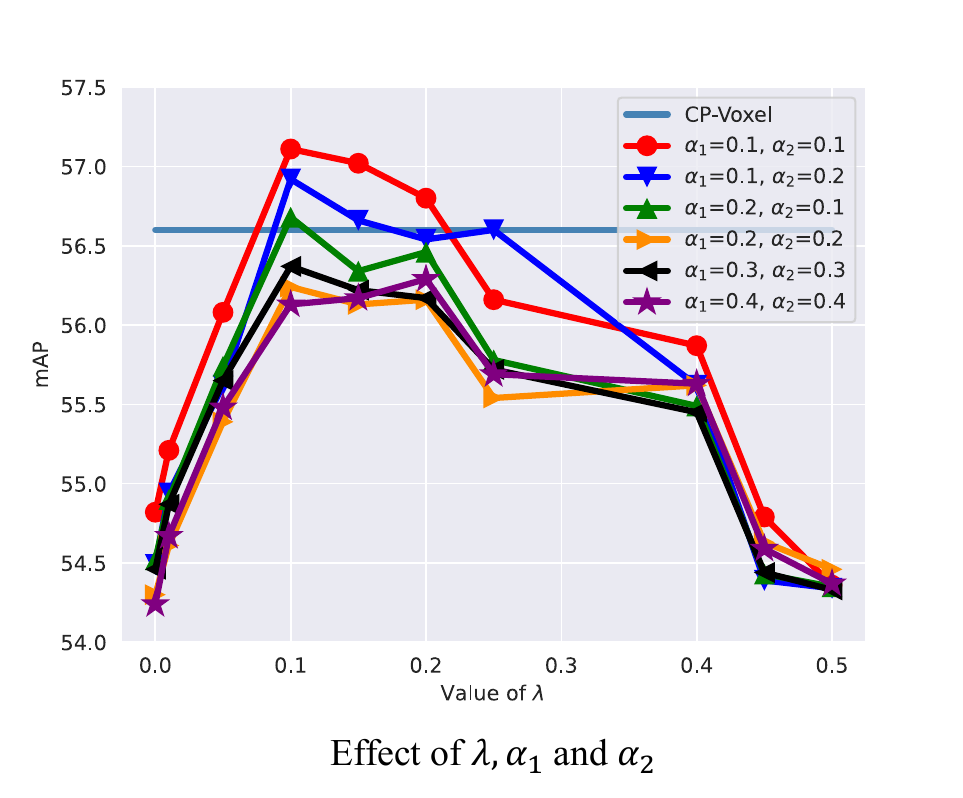}\\
\caption{Influence of $\lambda$, $\alpha_1$ and $\alpha_1$ using CP-Voxel-S~\cite{yang2022towards} on nuScenes~\cite{caesar2020nuscenes}.}
\label{fig:hyper}
\vspace{2mm}
\end{figure}

\begin{table}[]
\centering
\caption{Effects of different components in RDD with CP-Voxel-S and CP-Pillar-v0.4~\cite{yang2022towards} on nuScenes~\cite{caesar2020nuscenes}. All experiments use $\ell_2$ loss for distillation. Teacher models are marked in grey.}
\vspace{4mm}
\begin{tabular}{cccc}
\toprule
Model                           & \begin{tabular}[c]{@{}c@{}}Region\\ Selection\end{tabular} & mAP            & NDS            \\ \hline
\rowcolor[HTML]{EFEFEF} 
CP-Voxel                        & -                                                          & 56.6          & 64.7          \\
\multirow{4}{*}{CP-Voxel-S}     & -                                                          & 54.0          & 62.1          \\
& Hint                                                       & 52.9          & 61.1          \\
& FG                                                         & 54.3          & 62.6          \\
& \textbf{RDD}                                               & \textbf{57.1} & \textbf{65.0} \\ \hline
\rowcolor[HTML]{EFEFEF} 
CP-Pillar                       & -                                                          & 49.1          & 59.7          \\
\multirow{4}{*}{CP-Pillar-v0.4} & -                                                          & 46.5          & 55.5          \\
& Hint                                                       & 45.6          & 55.1          \\
& FG                                                         & 47.6          & 57.2          \\
& \textbf{RDD}                                               & \textbf{50.0} & \textbf{58.9} \\ \bottomrule
\end{tabular}
\label{tab:ablation_select}
\end{table}

For CP-Voxel and CP-Pillar~\cite{yin2021center}, we follow~\cite{yang2022towards} to adopt width and input resolution compression. Specifically, we compress the channels of $\{$encoder, backbone$\&$neck, head$\}$ ({\em i.e.}, $\{$Pillar Feature Encoding (PFE), Bird eye’s view Feature Encoding (BFE), Head$\}$ in~\cite{yang2022towards}) into $\{$1$\times$, 0.5$\times$, 0.5$\times\}$, $\{$0.75$\times$, 0.5$\times$, 0.5$\times\}$ and $\{$0.5$\times$, 0.25$\times$, 0.25$\times\}$, forming the CP-Voxel-S, CP-Voxel-XS and CP-Voxel-XXS. We follow~\cite{yang2022towards} to change the voxel size of CP-Pillar from 0.32 to 0.4, 0.48 and 0.64, forming the CP-Pillar-v0.4, CP-Pillar-v0.48 and CP-Pillar-v0.64. For SECOND~\cite{yan2018second}, we also follow~\cite{yang2022towards} to compress the width of SECOND~\cite{yan2018second}, where the channels of $\{$encoder, backbone$\&$neck$\}$ ({\em i.e.}, $\{$PFE, BFE$\}$ in~\cite{yang2022towards}) are reduced to $\{$0.75$\times$, 0.5$\times\}$ and $\{$0.5$\times$, 0.5$\times\}$,  forming the SECOND-S and SECOND-XS. For PointPillars~\cite{lang2019pointpillars}, we compress the width of PointPillars~\cite{lang2019pointpillars} forming the PointPillars-S and PointPillars-XS, in which the channels of $\{$encoder, backbone$\&$neck$\}$ are reduced to $\{$0.75$\times$, 0.5$\times\}$ and $\{$0.5$\times$, 0.5$\times\}$.

\begin{table}[]
\centering
\caption{Evaluating the components of RDD based on CP-Voxel-S and CP-Pillar-v0.4. RDD-F and RDD-L denote w/ or w/o distillation loss in Eq.\,(\ref{eq:feature_rdd}) and Eq.\,(\ref{eq:logit_rdd}), respectively.} 
\vspace{4mm}
\begin{tabular}{ccccc}
\toprule
Student                         & RDD-F        & RDD-L        & mAP            & NDS            \\ \hline
\multirow{4}{*}{CP-Voxel-S}     & -            & -            & 54.0          & 62.1          \\ \cline{2-5} 
& $\checkmark$ &              & 56.8          & 64.0          \\ 
&              & $\checkmark$ & 57.0          & 64.3          \\ 
& $\checkmark$ & $\checkmark$ & \textbf{57.1} & \textbf{65.0} \\ \hline
\multirow{4}{*}{CP-Pillar-v0.4} & -            & -            & 46.5          & 55.5          \\ \cline{2-5} 
& $\checkmark$ &              & 49.7          & 58.0          \\ 
&              & $\checkmark$ & 49.2          & 57.9          \\ 
& $\checkmark$ & $\checkmark$ & \textbf{50.0} & \textbf{58.9} \\ \bottomrule
\end{tabular}
\label{tab:ablation_loss}
\end{table}

\subsection{Ablation Study}
\label{sec:ablation}

\textbf{Hyper-parameter Selection}.
We first select hyper-parameters $\lambda$ of Eq.\,(\ref{eq:select_obj}) and $\alpha_1$, $\alpha_2$ of Eq.(\ref{eq:train_objective}) in this part, with experiments conducted on nuScenes dataset. Note that $\gamma$ in Eq.(\ref{eq:train_objective}) is sett following CenterPoint~\cite{yin2021center}, SECOND~\cite{yan2018second}, and Pointpillar~\cite{lang2019pointpillars} when conducted our expeirments on these detectors. We show the model performance (mAP) with different setups of hyper-parameters $\lambda$ in Fig.\,\ref{fig:hyper}, in which the performances increase first and then decrease with the uplift of $\lambda$ from left to right. Since $\lambda$ controls the decline of $\|{\bf m}\|_1$ related to the proportion of selected distillation-desired region pairs, $\lambda = 0$ denotes all the alternative region pairs are equally distilled leading to a unsatisfactory performance. While with $\lambda$ increasing to more than 0.1, the proportion of selected distillation-desired region pairs is not large enough for effectively eliminating the representation disparity, which also affects the performance. In conclusion, the CP-Voxel-S~\cite{yang2022towards} obtains better performances with $\lambda$ set as 0.1. The figure also shows that the CP-Voxel-S~\cite{yang2022towards} obtains best performances with $\alpha_1$ and $\alpha_2$ set as 0.2 and 0.2. However, enlarging $\alpha_1$ and $\alpha_2$ degenerates the performance of the detectors. Based on the ablative study above, we set hyper-parameters $\{\lambda, \alpha_1, \alpha_2\}$ as $\{$0.1, 0.2, 0.2$\}$ for the experiments in this paper.

\textbf{Component Ablation}.
We first compare our representation disparity-aware (RDD) region selecting method with other methods to select regions: Hint~\cite{chen2017learning} (using the neck feature without region mask) and FG~\cite{wang2019distilling}. We show the effectiveness of RDD with center-based 3D detectores~\cite{yin2021center} on nuScenes dataset~\cite{caesar2020nuscenes} in Tab.\,\ref{tab:ablation_select}. On the CP-Voxel-S~\cite{yang2022towards}, the introducing of RDD achieves improvements of the mAP and NDS by $\{$ 3.1\%, 4.2\%, 2.8\%$\}$ and $\{$2.9\%, 3.9\%, 2.4\%$\}$ compared to non-distillation, Hint~\cite{chen2017learning}, and FGFI~\cite{wang2019distilling}, under the same student-teacher framework. We also evaluate the RDD on the Pillar-based 3D detector, {\em i.e.}, CP-Pillar-v0.4~\cite{yang2022towards}. Our RDD selection method improves the mAP and NDS by $\{$3.5\%, 4.4\%, 2.4\%$\}$ and $\{$3.4\%, 3.8\%, 1.7\%$\}$ compared to non-distillation, Hint~\cite{chen2017learning}, and FGFI~\cite{wang2019distilling}, supervised by the same teacher (CP-Pillar~\cite{yin2021center}).

Then we evaluate the proposed distillation losses, {\em i.e.}, $\mathcal{L}_{feat}$ in Eq.\,(\ref{eq:feature_rdd}) (RDD-F) and $\mathcal{L}_{logit}$ in Eq.\,(\ref{eq:logit_rdd}) (RDD-L), in Tab.\,\ref{tab:ablation_loss}. As listed, RDD-F and RDD-L improve the performance when used alone, and the two losses further boost the performance considerably when combined together. For example, the RDD-F improve the mAP of CP-Voxel-S~\cite{yang2022towards} by 2.8\% and the RDD-L achieves 3.0\% mAP improvement. While combining the RDD-F and RDD-L together, the performance improvement achieves 3.1\%. These ablative experiments further validates the effectiveness of our method.

\begin{table}[]
\centering
\small
\caption{Experimental results on nuScenes~\cite{caesar2020nuscenes}. $\#$ F and $\#$ P indicate float operations (FLOPs) and parameters of the detector. 
Teacher models are marked in gray shadow.}
\vspace{4mm}
\begin{tabular}{ccccc}
\hline
Detector                                                                     & \begin{tabular}[c]{@{}c@{}}$\#$ F/$\#$ P\\ (G/M)\end{tabular}   & Method           & mAP ($\uparrow$) & NDS ($\uparrow$) \\ \hline
\rowcolor[HTML]{EFEFEF} 
CP-Voxel                                                                     & 114.8 / 7.8                                                     & -            & 56.6            & 64.7            \\
&                                                                 & No Distill   & 54.0            & 62.1            \\ \cdashline{3-5}
&                                                                 & GID-L        & 54.1            & 62.1            \\
&                                                                 & GID-F        & 53.4            & 61.8            \\
&                                                                 & FG           & 54.3            & 62.6            \\
&                                                                 & LAD          & 53.0            & 61.7            \\ \cdashline{3-5}
&                                                                 & PP Logit KD  & 55.7            & 64.3            \\
\multirow{-7}{*}{\begin{tabular}[c]{@{}c@{}}CP-Voxel\\ -S\end{tabular}}      & \multirow{-7}{*}{47.8 / 4.0}                                    & \textbf{RDD} & \textbf{57.1}   & \textbf{65.0}   \\ \hdashline
&                                                                 & No Distill   & 53.0            & 61.8            \\ \cdashline{3-5}
&                                                                 & GID-L        & 53.2            & 61.6            \\
&                                                                 & GID-F        & 52.9            & 61.3            \\
&                                                                 & FG           & 53.3            & 61.8            \\
&                                                                 & LAD          & 53.0            & 61.5           \\ \cdashline{3-5}
&                                                                 & PP Logit KD  & 53.5            & 61.7            \\
\multirow{-7}{*}{\begin{tabular}[c]{@{}c@{}}CP-Voxel\\ -XS\end{tabular}}     & \multirow{-7}{*}{36.0 / 2.8}                                    & \textbf{RDD} & \textbf{54.0}   & \textbf{62.1}   \\ \hdashline
&                                                                 & No Distill   & 46.7            & 55.5            \\ \cdashline{3-5}
&                                                                 & GID-L        & 46.8            & 56.5            \\
&                                                                 & GID-F        & 47.0            & 56.4            \\
&                                                                 & FG           & 47.1            & 56.2            \\
&                                                                 & LAD          & 46.6            & 55.3            \\ \cdashline{3-5}
&                                                                 & PP Logit KD  & 47.9            & 57.8            \\
\multirow{-7}{*}{\begin{tabular}[c]{@{}c@{}}CP-Voxel\\ -XXS\end{tabular}}    & \multirow{-7}{*}{12.0 / 1.0}                                    & \textbf{RDD} & \textbf{49.4}   & \textbf{57.9}   \\ \hline
\rowcolor[HTML]{EFEFEF} 
CP-Pillar                                                                    & 333.9 / 5.2                                                     & -            & 49.1            & 59.7            \\ 
&                                                                 & No Distill   & 46.5            & 55.5            \\ \cdashline{3-5}
&                                                                 & GID-L        & 47.3            & 56.4            \\
&                                                                 & GID-F        & 47.6            & 56.8            \\
&                                                                 & FG           & 47.7            & 57.2            \\
&                                                                 & LAD          & 46.9            & 55.7            \\ \cdashline{3-5}
&                                                                 & PP Logit KD  & 48.6            & 57.5            \\
\multirow{-7}{*}{\begin{tabular}[c]{@{}c@{}}CP-Pillar\\ -v0.4\end{tabular}}  & \multirow{-7}{*}{212.9 / 5.2}                                   & \textbf{RDD} & \textbf{50.0}   & \textbf{58.9}   \\ \hdashline
&                                                                 & No Distill   & 45.3            & 54.2            \\ \cdashline{3-5}
&                                                                 & GID-L        & 45.4            & 54.5            \\
&                                                                 & GID-F        & 46.1            & 55.3            \\
&                                                                 & FG           & 47.0            & 56.2            \\
&                                                                 & LAD          & 45.2            & 54.4            \\ \cdashline{3-5}
&                                                                 & PP Logit KD  & 47.3            & 57.5            \\
\multirow{-7}{*}{\begin{tabular}[c]{@{}c@{}}CP-Pillar\\ -v0.48\end{tabular}} & \multirow{-7}{*}{149.4 / 5.2}                                   & \textbf{RDD} & \textbf{48.8}   & \textbf{58.5}   \\ \hdashline
&                                                                 & No Distill   & 44.0            & 52.3            \\ \cdashline{3-5}
&                                                                 & GID-L        & 44.2            & 52.6            \\
&                                                                 & GID-F        & 44.4            & 53.9            \\
&                                                                 & FG           & 44.7            & 53.7            \\
&                                                                 & LAD          & 43.9            & 53.2            \\ \cdashline{3-5}
&                                                                 & PP Logit KD  & 45.0            & 55.9            \\
\multirow{-7}{*}{\begin{tabular}[c]{@{}c@{}}CP-Pillar\\ -v0.64\end{tabular}} & \multirow{-7}{*}{85.1 / 5.2}                                    & \textbf{RDD} & \textbf{45.8}   & \textbf{56.1}        \\ \hline
\end{tabular}
\label{tab:nuscenes}
\end{table}

\subsection{Results on nuScenes}
\label{sec:nuScenes}
We first compare our method with image-based 2D and 3D distillation methods for the task of 3D object detection on nuScenes~\cite{caesar2020nuscenes}. Note that for width compressed students, we use the same input voxel size as~\cite{yin2021center}, {\em i.e.} $0.1$. We mainly discuss the mAP and NDS (default nuScenes metric) in the following. We evaluate the proposed RDD on CenterPoint detectors~\cite{yin2021center} in Tab.\,\ref{tab:nuscenes}. For CP-Voxel-S, compared to non-distillation of GID-L~\cite{dai2021general}, GID-F~\cite{dai2021general}, FG~\cite{wang2019distilling} and LAD~\cite{nguyen2022improving}, our RDD boosts the performance of mAP and NDS by $\{$3.1\%, 3.0\%, 3.7\%, 2.8\%, 4.1\%$\}$ and $\{$2.9\%, 2.9\%, 3.2\%, 2.4\%, 3.3\%$\}$, respectively. Moreover, our RDD improves the mAP and NDS of CP-Voxel-S by $1.4\%$ and $0.7\%$ respectively, compared with previous state-of-the-art 3D distillation method (PP-logit-KD). It is worth noting that our RDD trained CP-Voxel-S even surpasses the teacher detectors by 0.5\% mAP and 0.3\% NDS but taking up only 41.6\% FLOPs and 51.3\% parameters of the teacher, which is a significant achievement. For CP-Voxel-XS, our RDD improves the performance of mAP and NDS by $\{$1.0\%, 0.8\%, 1.1\%, 0.7\%, 1.0\%$\}$ and $\{$0.3\%, 0.5\%, 0.8\%, 0.3\%, 0.6\%$\}$, compared to non-distillation of GID-L~\cite{dai2021general}, GID-F~\cite{dai2021general}, FG~\cite{wang2019distilling} and LAD~\cite{nguyen2022improving}. In addition, our RDD improves the mAP and NDS of CP-Voxel-XS by $0.5\%$ and $0.4\%$, compared with previous state-of-the-art 3D distillation method (PP Logit KD). For CP-Voxel-XXS, our RDD surpasses non-distillation of GID-L~\cite{dai2021general}, GID-F~\cite{dai2021general}, FG~\cite{wang2019distilling} and LAD~\cite{nguyen2022improving} by $\{$2.7\%, 2.6\%, 2.4\%, 2.3\%, 2.8\%$\}$ mAP and $\{$2.4\%, 1.4\%, 1.5\%, 1.7\%, 1.6\%$\}$ NDS. Moreover, our RDD boosts the mAP and NDS of CP-Voxel-XXS by $1.5\%$ and $0.1\%$ respectively, compared with previous state-of-the-art 3D distillation method (PP Logit KD). Above experiments well validates the effectiveness of our method. 

Besides, our method generates convincing results on CP-Pillar based detectors~\cite{yin2021center}. As shown in the 24-th to 45-th rows of Tab.\,\ref{tab:nuscenes}, the performance of the proposed RDD with CP-Pillar-v0.4, CP-Pillar-v0.48 and CP-Pillar-v0.64 outperforms the non-distillation baseline by $\{$3.5\%, 3.5\%, 1.8\%$\}$ and $\{$3.4\%, 4.3\%, 3.8\%$\}$ on mAP and NDS, a large margin. Compared with previous state-of-the-art 3D distillation methods, our RDD achieves $\{$1.4\%, 1.5\%, 0.8\%$\}$ and $\{$1.4\%, 1.0\%, 0.2\%$\}$ improvement on mAP and NAS respectively with CP-Pillar-v0.4, CP-Pillar-v0.48 and CP-Pillar-v0.64, which well validates the efficacy of our method. The above experimental results prove the superiority of our RDD method on both Voxel-based and Pillar-based CenterPoint~\cite{yin2021center}.

\begin{table}[]
\centering
\small
\caption{Experimental results  for 3D detection on KITTI~\cite{geiger2012we}. $\#$ F and $\#$ P indicate float operations (FLOPs) and parameters of the detector. 
Teacher models are marked in gray shadow.}
\vspace{4mm}
\begin{tabular}{ccccc}
\toprule
&                                                                        &                                                                        &                      & 3D                                                         \\ \cline{5-5} 
\multirow{-2}{*}{Detector}       & \multirow{-2}{*}{\begin{tabular}[c]{@{}c@{}}$\#$ F\\ (G)\end{tabular}} & \multirow{-2}{*}{\begin{tabular}[c]{@{}c@{}}$\#$ P\\ (M)\end{tabular}} & \multirow{-2}{*}{Method} & \begin{tabular}[c]{@{}c@{}}Moderate\\ mAP@R40($\uparrow$)\end{tabular} \\ \hline
\rowcolor[HTML]{EFEFEF} 
SECOND                           & 80.5                                                                   & 5.3                                                                    & -                    & 67.2                                                      \\ 
&                                                                        &                                                                        & No Distill           & 65.6                                                      \\ \cdashline{4-5}
&                                                                        &                                                                        & GID-L                & 66.3                                                      \\
&                                                                        &                                                                        & GID-F                & 66.8                                                      \\
&                                                                        &                                                                        & FG                   & 66.6                                                      \\
&                                                                        &                                                                        & LAD                  & 67.0                                                      \\ \cdashline{4-5}
&                                                                        &                                                                        &PointDistiller        & 67.8                                                      \\
&                                                                        &                                                                        & PP Logit KD          & 67.7                                                      \\
\multirow{-7}{*}{\begin{tabular}[c]{@{}c@{}}SECOND\\ -S\end{tabular}}      
& \multirow{-7}{*}{23.0}                                                 & \multirow{-7}{*}{1.6}                                                  & \textbf{RDD}         & \textbf{68.2}                                             \\ \hdashline
&                                                                        &                                                                        & No Distill           & 64.2                                                      \\ \cdashline{4-5}
&                                                                        &                                                                        & GID-L                & 65.0                                                      \\
&                                                                        &                                                                        & GID-F                & 65.2                                                      \\
&                                                                        &                                                                        & FG                   & 65.3                                                      \\
&                                                                        &                                                                        & LAD                  & 65.7                                                      \\ \cdashline{4-5}
&                                                                        &                                                                        &PointDistiller        & 66.3                                                      \\
&                                                                        &                                                                        & PP Logit KD          & 66.4                                                      \\
\multirow{-7}{*}{\begin{tabular}[c]{@{}c@{}}SECOND\\ -XS\end{tabular}}      
& \multirow{-7}{*}{20.5}                                                 & \multirow{-7}{*}{1.4}                                                  & \textbf{RDD}         & \textbf{67.0}                                             \\ \hline
\rowcolor[HTML]{EFEFEF} 
PointPillars                     & 34.3                                                                   & 4.8                                                                    & -                    & 60.3                                                       \\ 
&                                                                        &                                                                        & No Distill           & 58.6                                                      \\ \cdashline{4-5}
&                                                                        &                                                                        & GID-L                & 58.9                                                      \\
&                                                                        &                                                                        & GID-F                & 59.1                                                      \\
&                                                                        &                                                                        & FG                   & 59.4                                                      \\
&                                                                        &                                                                        & LAD                  & 59.2                                                      \\ \cdashline{4-5}
&                                                                        &                                                                        &PointDistiller        & 62.3                                                      \\
&                                                                        &                                                                        & PP Logit KD          & 62.2                                                      \\
\multirow{-7}{*}{\begin{tabular}[c]{@{}c@{}}PointPillars\\ -S\end{tabular}}      
& \multirow{-7}{*}{9.8}                                                  & \multirow{-7}{*}{1.5}                                                  & \textbf{RDD}         &  \textbf{63.0}                                            \\ \hdashline
&                                                                        &                                                                        & No Distill           & 58.9                                                      \\ \cdashline{4-5}
&                                                                        &                                                                        & GID-L                & 59.2                                                      \\
&                                                                        &                                                                        & GID-F                & 59.6                                                      \\
&                                                                        &                                                                        & FG                   & 59.4                                                      \\
&                                                                        &                                                                        & LAD                  & 59.7                                                      \\ \cdashline{4-5}
&                                                                        &                                                                        &PointDistiller        & 60.0                                                      \\
&                                                                        &                                                                        & PP Logit KD          & 60.2                                                      \\
\multirow{-7}{*}{\begin{tabular}[c]{@{}c@{}}PointPillars\\ -XS\end{tabular}}      
& \multirow{-7}{*}{8.7}                                                  & \multirow{-7}{*}{1.3}                                                  & \textbf{RDD}         & \textbf{60.9}                                             \\ \bottomrule

\end{tabular}
\label{tab:kitti}
\end{table}

\subsection{Results on KITTI}
\label{sec:kitti}
We further show that our RDD can generalize well to KITTI~\cite{geiger2012we} dataset with anchor-based detectors SECOND~\cite{yan2018second} and PointPillar~\cite{lang2019pointpillars}. As shown in Tab.\,\ref{tab:kitti}, for SECOND-S, compared to non-distillation, GID-L~\cite{dai2021general}, GID-F~\cite{dai2021general}, FG~\cite{wang2019distilling} and LAD~\cite{nguyen2022improving}, our RDD boosts the performance of moderate mAP@R40 by $\{$2.6\%, 1.9\%, 1.4\%, 1.6\%, 1.2\%$\}$. Moreover, our RDD improves the moderate mAP@R40 of SECOND-S by 0.4\% and 0.5\%, compared with PointDistiller~\cite{zhang2022pointdistiller} and PP Logit KD~\cite{yang2022towards}. And SECOND-S trained with our RDD even surpasses its teacher model by 1.0\%. And for SECOND-XS, our RDD surpasses previous image-based 2D distillation methods GID-L~\cite{dai2021general}, GID-F~\cite{dai2021general}, FG~\cite{wang2019distilling} and LAD~\cite{nguyen2022improving} by by $\{$2.8\%, 2.0\%, 1.8\%, 1.7\%, 1.3\%$\}$ in moderate mAP@R40. SECOND-XS trained with our RDD also surpasses previous 3D distillation method, PoinrDistiller and PP Logit KD, by  0.7\% and 0.6\% on the moderate mAP@R40, which well validates the effectiveness of our method. Moreover, our method can also be generalized to PointPillars~\cite{lang2019pointpillars}. As shown in the 20-th to 36-th rows of Tab.\,\ref{tab:kitti}, the performance of the proposed RDD with PointPillars-S and PointPillars-XS outperforms the non-distillation baseline by 4.5\% and 2.0\% on the moderate mAP@R40, a large margin. Compared with PoinrDistiller, our RDD achieves 1.7\% and 0.9\% improvement on the moderate mAP@R40 with PointPillars-S and PointPillars-XS. Our RDD also surpasses PP Logit KD by 1.8\% and 0.7\% improvement on the moderate mAP@R40 with PointPillars-S and PointPillars-XS, which well validates the efficacy of our method. The above results are of great significance in the 3D LiDAR-based object detection.


\section{Conclusion}
\label{sec:conclusion}
This paper presents a novel method for training compact 3D LiDAR-based detectors with knowledge distillation to eliminate the representation disparity (RDD). RDD employs a information bottleneck (IB) principle to select the regions with maximum representation disparity and proposes effective distillation losses to supervise the representation disparity. As a result, our RDD significantly boosts the performance of compact 3D detectors. Extensive experiments show that RDD surpasses state-of-the-art compact 3D detectors and other knowledge distillation methods in 3D LiDAR-based object detection.

\section{Acknowledgement}
This work was supported in part by the National Natural Science Foundation of China under Grant 62076016, under Grant 61827901, “One Thousand Plan” projects in Jiangxi Province Jxsg2023102268, Foundation of China Energy Project GJNY-19-90.

{\small
\bibliographystyle{ieee_fullname}
\bibliography{egbib}
}

\end{document}